\documentclass[conference]{IEEEtran}
\IEEEoverridecommandlockouts
\usepackage{cite}
\usepackage{amsmath,amssymb,amsfonts}
\usepackage{algorithmic}
\usepackage{graphicx}
\graphicspath{ {Images/} }
\usepackage{cleveref}
\usepackage{textcomp}
\usepackage{xcolor}
\usepackage{enumitem}
\def\BibTeX{{\rm B\kern-.05em{\sc i\kern-.025em b}\kern-.08em
    T\kern-.1667em\lower.7ex\hbox{E}\kern-.125emX}}
\begin{document}

\title{Comparing Direct and Indirect Representations for Environment-Specific Robot Component Design}

\author{Jack Collins$^{1,2}$, Ben Cottier$^{1,3}$, and David Howard$^{1}$  % <-this % stops a space
% \thanks{*This work was not supported by any organization}% <-this % stops a space
\thanks{$^{1}$with Data61, Commonwealth Scientific and Industrial Research Organisation (CSIRO), Brisbane, Australia
        }%
\thanks{$^{2}$with the Queensland University of Technology, Brisbane, Australia
        }%
\thanks{$^{3}$with the University of Queensland, Brisbane, Australia
        }%
}

\maketitle

\begin{abstract}
We compare two representations used to define the morphology of legs for a hexapod robot, which are subsequently 3D printed.  A leg morphology occupies a set of voxels in a voxel grid. One method, a direct representation, uses a collection of Bezier splines. The second, an indirect method, utilises CPPN-NEAT.  In our first experiment,  we investigate two strategies to post-process the CPPN output and ensure leg length constraints are met. The first uses an adaptive threshold on the output neuron, the second, previously reported in the literature, scales the largest generated artefact to our desired length. In our second experiment, we build on our past work that evolves the tibia of a hexapod to provide environment-specific performance benefits. 
We compare the performance of our direct and indirect legs across three distinct environments, represented in a high-fidelity simulator. Results are significant and support our hypothesis that the indirect representation allows for further exploration of the design space leading to improved fitness. 

% Main Points:
% -CPPN
% -NEAT
% -Bezier
% -Thresholding
% -Scaling
% -3D printing
% -evolution of targeted components

\end{abstract}

\section{Introduction}
Development of robotic platforms involves expert knowledge applied to high-dimensional problems with solutions constrained by human preconceptions and our inability to solve problems with vast search spaces. Past research into Evolutionary Robotics (ER) has demonstrated the functionality of evolved designs \cite{Lipson2000AutomaticLifeforms,Funes1998EvolutionaryRobots} through the automatic creation of robot components that offer heightened environment-specific performance. Reducing human input by automating Computer Aided Design (CAD) of robots allows for a larger unbiased design spaces to be explored. Unlike traditional ER, which to date has struggled to match the performance of traditionally-designed robots, our approach \cite{Collins2018TowardsComponents} reduces the domain of the problem by focusing only on the performance critical components of a robot, in this case the legs of a hexapod robot. Our approach specifically permits the easy instantiation of optimised components using a standard 3D file format and 3D printer; the resulting robot displays environment-tailored performance whilst being able to perform multiple mission types using a well-developed control stack.

The design space for evolving artefacts can be simplified to a voxelised grid with predefined size constraints and resolution \cite{Clune2011EvolvingBiology}. As a proof-of-concept our previous experiments used a direct representation to define the occupancy of the grid. The representation took the form of a collection of 3D Bezier splines dictating the ``occupied" voxels with the number, thickness and control points of the Beziers acting as evolvable parameters. The Bezier representation allows for easy mapping between genotype and phenotype, increasing the evolvability of the representation, however, the attainable regions in the design space are limited to those that can be expressed as collections of Bezier splines.  This motivates our research into indirect component representations.

The current state of the art uses Compositional Pattern Producing Networks (CPPN) \cite{Cheney2013UnshacklingEncoding}. CPPNS are essentially feedforward Multi-Layer Perceptron networks \cite{Stanley2007CompositionalDevelopment} augmented with an expanded set of activation functions.  Input neurons specify the spatial coordinates of the voxel under consideration, and a single output neuron defines the occupancy of the voxel (full of material, or empty).  CPPNs are frequently evolved using Neuroevolution of Augmenting Topologies (NEAT) \cite{Richards2016ShapeCPPNs}. This form of representation and evolution is described in literature as following a biological approach to defining morphology and therefore produces organic leg formations \cite{Stanley2007CompositionalDevelopment,Stanley2002EvolvingTopologies}. A common issue with CPPNs is that design constraints are not easy to enforce (although we note some attempts to remedy this \cite{Auerbach2014EnvironmentalMachines}).  As legs are required to span the vertical axis with contiguous voxels, we implement two methods for ensuring this constraint is satisfied. The first of which applies an adaptive threshold while the second implementation adopts a scaling approach to re-size undersized elements.

% Utilising a high-fidelity simulator and three distinct simulation environments we evolve the morphology of hexapod tibia to be high performing in each environment. The search for legs evolved using both representations is parallelised across the nodes of a high-performance compute cluster (HPC), providing timely results. We pose the hypothesis that \emph{indirect representations allow for greater exploration of the environment, in turn gaining a higher fitness}. To test this hypothesis we formulate the following research questions:

Utilising a high-fidelity simulator and three distinct simulation environments we evolve the morphology of hexapod tibia to be high performing in each environment. The search for legs evolved using both representations is parallelised across the nodes of a high-performance compute cluster (HPC), providing timely results. Evolved legs can easily be instantiated in reality with a bracket using the standard .STL file format; the printed bracket can mount directly onto a number of CSIRO hexapod platforms. 

We pose the hypothesis that \emph{indirect representations allow for greater exploration of the environment, in turn gaining a higher fitness}. To test this hypothesis we formulate the following research questions:

\begin{enumerate}
    \item Which of the two methods engineered for ensuring compliance of the CPPN provides more optimal solutions?
    \item Are legs evolved via CPPN-NEAT awarded increased fitness over those evolved using a direct encoding?
    \item Do indirect encodings further exploit the environment when compared to direct encodings?
\end{enumerate}

Results are statistically significant and show a large difference between the two representations, proving the validity of our hypothesis. Analysis of CPPN results show scaling the CPPN output affords a greater mean fitness than that of a leg population evolved using CPPNs and an adaptive threshold. Furthermore, when a scaled, indirect representation leg contends against a direct, Bezier representation the indirect legs are proven to be higher performing. Future research will investigate these findings and the transferability of our approach using a test rig built to replicate our simulation scene.

% By using a highly realistic simulator at the expense of increased simulation times we look to reduce the reality gap. 

% NEXT STEP IS DOING IT IN REALITY WITH A TEST RIG.  HOWEVER NOTE THAT REALITY GAP MAY BE LARGELY UNAFFECTED; WHY?

% WE HAVE SHOWN XX YY

% NOW WE WANT TO REPLACE WITH STATE OF THE ART INDRIRECT ENCODING.
% SAY DIRECT VS INDIRECT, THEN freedom vs searchability...

% ADD NEW PICS
% ADD SAMPLE LEG PICS FROM EACH
% NEXT STEP IS DOING IT IN REALITY WITH A TEST RIG.  HOWEVER NOTE THAT REALITY GAP MAY BE LARGELY UNAFFECTED; WHY?

% FIRST WE COMPARE STANDARD TO A DIFFENT WAY BASED ON ADAPTIVE THRESHOLD, THEN USE BEST FOR THE MAIN EXPERIMENT, IN WHICH WE...
% SECOND, WE ADDRESS THE ISSUE OF MAKING CPPN OUTPUT SUITABLE...adaptive threshold vs. rescale to voxel thing.

% WE HAVE PREVIOUSLY SHOWN!

\section{Related Research}

\subsection{Evolutionary Robotics}

ER research looks to create robot morphologies and/or controllers that adapt across generations to environmental pressures; a synthetic resemblance of evolution in the real world. To evolve robot morphologies evolutionary methods require an encoding that can be representative of a genome, much like our DNA acts as the genotype for our physical phenotype. Each representation, together with the evolutionary operators that manipulate it, sits somewhere on a \emph{continuum of reality} that trades off between phenotypic complexity and evolvability. At one extreme, we see direct, integer-based or binary representations, heavily abstracted from natural reality and with limited phenotypic complexity, yet typically highly evolvable. Indirect representations are more scalable, and so offer heightened complexity, yet because of the indirection between genotype and phenotype, can be difficult to evolve.

\begin{figure}[tb!]
  \centering
  \includegraphics[width=\linewidth]{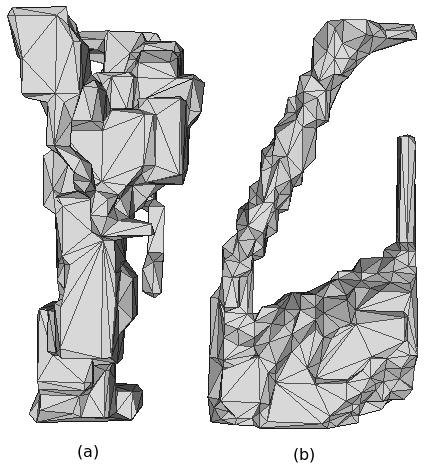}
  \caption{(a) A leg evolved using a direct representation from a collection of Bezier splines and an Evolutionary Algorithm; (b) A leg developed using an indirect representation from a Compositional Pattern Producing Network paired with Neuroevolution of Augmenting Topologies Evolutionary Algorithm. Both legs were evolved to be high-performing in a simulated gravel environment.}
  \label{GravelCompare}
\end{figure}

It is intuitive to believe that indirect representations provide performance benefits over direct encodings, however, past research has shown that this isn't necessarily the case \cite{CluneJeffandOfria2008HowDecreases}. Balancing evolvability and expressiveness appears to be problem dependant. Because of this, this paper compares our previous direct representation to an indirect alternative based on CPPN-NEAT. Figure \ref{GravelCompare} visually demonstrates the differences between two legs evolved using direct and indirect methods.

\subsection{Shape Optimisation}

An established goal of ER is to produce robots with evolved morphologies that are human-competitive. External from ER, shape optimisation paired with evolutionary methods has already produced human-competitive results. We have seen evolved antenna \cite{Lohn2005AnMission}, truss design \cite{Fenton2014AutomaticEvolution} and lens system design \cite{Gagne2008Human-competitiveStrategies}.

Our approach borrows from shape optimisation in that we evolve physically static components of the robot. In past research we have seen the automated creation of CAD files for jewellery \cite{Wannarumon2008AestheticDesign}, 3D art \cite{Clune2013UploadEvolution} and more functional designs including wind turbine blades \cite{Preen2015TowardTurbines}. Representations for this area of work include direct encodings with current trends towards generative representations. By applying concepts from this body of work we reduce the problem dimenionality of robot morphology evolution.

\subsection{Compositional Pattern Producing Networks}
CPPN's introduced by Stanley \cite{Stanley2007CompositionalDevelopment}, abstract the process of natural development to provide a solution to indirectly mapping genotype to phenotype. The product mimics biology without the need to simulate the computationally intensive, low-level interactions of chemicals and proteins \cite{Clune2011EvolvingBiology}.

A range of functions are used by a CPPN to produce biological features whilst ANNs are commonly restricted to either sigmoid or hyperbolic tangent functions \cite{Stanley2009ANetworks}. Different activation functions allow for different characteristics to be displayed, for example a Gaussian function can provide bilateral symmetry to the output, while periodic functions like sine and cosine can produce repetition and segmentation in the end product \cite{Stanley2006ExploitingDevelopment}. A noteworthy feature of CPPN's is their ability to be re-queried to define morphologies to higher resolution without the need for the network to be re-evolved \cite{Auerbach2010EvolvingStructures}. 

Since their inception CPPNs have been applied to many problems requiring the physical instantiation of evolved artefacts. Classes of problems stretch between structural beam design \cite{Richards2016ShapeCPPNs} through to computational creativity \cite{Lehman2016CreativeEngines,Clune2011EvolvingBiology} with ER perceived early on as a possible application \cite{Auerbach2010EvolvingStructures,Auerbach2010DynamicControl}. Of note is the work by Secretan \emph{et al.} \cite{Secretan2008Picbreeder:Online} on Picbreeder, a system of evolving pictures collaboratively online and EndlessForms by Clune \emph{et al.} \cite{Clune2013UploadEvolution} where 3D objects in .STL format are interactively evolved online. 

Research into the use of CPPNs for ER has generated solutions to simulated tasks by evolving both the control and morphology of robots \cite{Auerbach2010DynamicControl,Auerbach2011EvolvingConnections,Auerbach2012OnRobots}. Notably Cheney \emph{et al.} \cite{Cheney2013UnshacklingEncoding} evolved CPPNs in their VoxCAD environment and generated life-like organisms in the form of soft robots. Further progression of this work has seen CPPNs and VoxCAD applied to terrestrial, aquatic and tight space environments as well as the evolutionary development of plants \cite{Corucci2018EvolvingTransitions,Corucci2016EvolvingCreatures,Cheney2015EvolvingSpaces,Corucci2017EvolutionaryPlants}.  We look to remove CPPN evolved robots from simulated environments and allow for instantiation in the real world whilst also increasing the size and resolution of the design space. 

\subsection{NEAT}
CPPN's share similar structure to ANNs, this allows them to utilise evolutionary methods first designed for ANNs.  Neuroevolution of Augmenting Topologies (NEAT) has been described in detail by Stanley \cite{Stanley2002EvolvingTopologies}, so we briefly summarise it here.  NEAT evolves the weights and topology of a network \cite{Stanley2009ANetworks}, and when paired with CPPNs also evolves the activation function. CPPN-NEAT allows for networks with increasing complexity to evolve with time, creating outputs with developing intricacies \cite{Stanley2009ANetworks}.

\section{Experimental Setup}
To promote reproducible results Appendix \ref{Appendix A} details the parameters used herein. Both experiments follow the methodology outlined below, and last for 50 generations with a population size of 20. Ten repeats of each experiment are carried out, and were statistically compared using the Mann-Whitney U-Test\footnote{Does not require normally-distributed samples} with a significance of $P<0.05$ .

Each experiment begins with a population of $20$ legs randomly initialised using one of two representations. The population is evaluated in simulation with each member assigned a fitness. Dependant on representation a new child population of legs is spawned using the evolutionary processes described below. Evaluation and creation of a child population repeats until termination after 50 generations.

% PARAGRAPH SAYING "TO PERFORM AN EXPERIMENT, WE RANDOMLY INITIALISE A POPULATION OF 20 LEGS USING A GIVEN REPRESENTATION AND ASCERTAIN THEIR FITNESS.  WE THEN GENERATE A CHILD POPULATION USING A REPRESENTATION-SPECIFIC EVOLUTIONARY PROCESS AS DESCRIBED BELOW, AND WHICH TERMINATES AFTER 100 GENERATIONS.

\subsection{Methodology}

Project Chrono\footnote{http://projectchrono.org} is a high-fidelity simulation environment capable of modelling complex environmental interactions through a discrete element method (DEM). It is able to simulate rigid body, flexible-body and fluid dynamics with consideration for frictions, foot-slip and torque limits. Legs are imported into the simulator as .OBJ files. Three environments are modelled in the simulator:
\begin{enumerate}[label=\Alph*]
    \item A deformable terrain model with characteristics of soil;
    \item A DEM environment with particles representing loose gravel;
    \item A DEM environment with fluid particles encoded with water-like properties.
\end{enumerate}

\begin{figure}
  \centering
  \includegraphics[width=\linewidth]{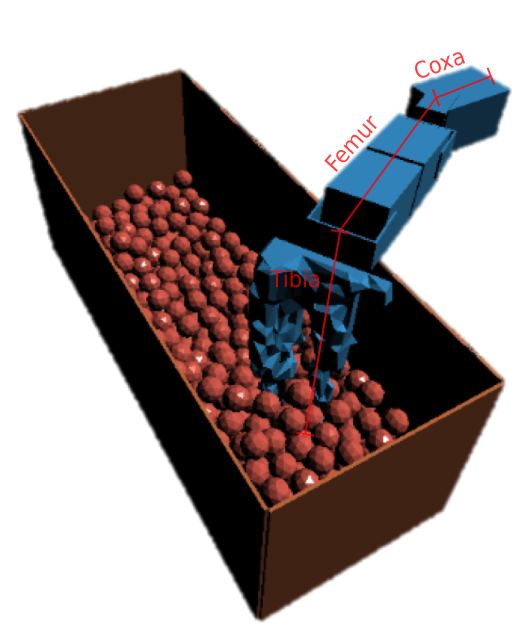}
  \caption{An example of the gravel environment where the tibia steps through the terrain within the bin. A step consists of a step-down phase, sweep across phase and a step-up phase. Leg anatomy is labelled with actuation of the coxa, femur and tibia joints.}
  \label{SimEnvironment}
\end{figure}

Figure \ref{SimEnvironment} shows a gravel simulation with elements including: a bin; terrain within the bin; three primitive motors with brackets (indicative of Dynamixel actuators); and evolved tibia. Project Chrono has the ability to record the torques required to complete motor rotations and we use this as part of the evaluation.

The torque used by each of the three motors at each time-step is summed whilst completing a predefined ``step" within the bin. The accumulated torque provides the first term in our fitness function. Intuitively, by minimising the torques required to walk through an environment we increase the attainable mission times.

A ``step" through the terrain is split into three phases: the step-down, sweep across and step-up phases. In the step down phase, the femur rotates from $+30^{\circ}$ to neutral while the tibia rotates from $-30^{\circ}$ to neutral.  In the second phase, the coxa sweeps $60^{\circ}$ backwards from $+30^{\circ}$; the step-up is the inverse of the step-down. 

Fitness is calculated according to \cref{eq:fitness}, where $\tau$ is the combined torque (Nm) of all three motors at each 1ms time step, $n_{steps}$ is the number of simulation steps (3000) and $\delta$ is the percentage of occupied voxels.  Larger values indicate higher performance.

\begin{equation}
    f = \displaystyle\frac{1}{\frac{\sum \tau}{n_{steps}}+\bigg(\frac{\sum \tau}{n_{steps}}\times \frac{\delta}{5}\bigg)}
    \label{eq:fitness}
\end{equation}

As we instantiate legs using a 3D printer, we wish to minimise superfluous material as it is time consuming, wasteful and expensive to print. To reduce unnecessary material we add a second term to our fitness function that penalises legs with a large number of occupied cells. The second term is of less significance than the first and as such is given a lessened effect on the final fitness.

An evaluation can take anywhere between 2-5mins to complete on the HPC, this can translate to environments taking between 3hrs (soil) and 7 days (fluid) to complete an evolutionary optimisation.

\subsection{Representations}

The design space for each leg is a discretised into a $16\times32\times16$ voxel grid with voxel resolution set to 5 millimetres. Each voxel can be in one of two states, ``full'' or ``empty'', with the occupancy defined by the representation. Once generated, a voxelised artefact is converted into .STL format and processed by Meshlab\footnote{http://www.meshlab.net/}. Meshlab processing scales, transforms, rotates and decimates the mesh. The processed mesh is saved as an .OBJ for direct import into the simulator. 

\subsubsection{CPPN-NEAT}
\begin{figure*}[thpb]
  \centering
  \includegraphics[width=\textwidth]{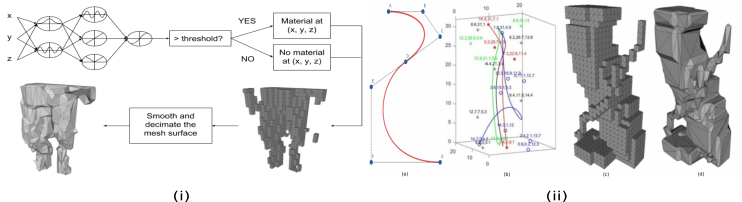}
  \caption{(i) All voxel locations are passed iteratively to the CPPN network as x,y,z inputs. The output is a float between 0 and 1 with a threshold deciding whether the voxel will contain material or not. A voxelised leg is then converted to an .STL file format and processed in Meshlab to produce an .OBJ simulatable file. (ii) (a) A 2D Bezier line made with 7 control points (b) The graphical representation of a leg using 3D Beziers. A leg can have between 5-10 splines, each with between 3 to 8 control points (c) Voxelised conversion of Bezier splines in .STL format (c) .STL with post-processing converted into .OBJ}
  \label{Representation}
\end{figure*}

Our CPPN-NEAT implementation is an indirect encoding for describing the occupancy of the voxelised space. Following the majority of past research using CPPN-NEAT the x,y and z positions of all the voxel locations are passed as parameters to the input nodes of the network and a float value between $0$ and $1$ is recieved from the output node. The float is converted to a ``full" or ``empty" voxel dependant on a set threshold, generally $0.5$. Figure \ref{Representation}(i) shows the generation process of a leg. 

The Evolutionary Algorithm (EA) for evolving the network, NEAT, evolves the structure, weights and activation functions of the CPPN. Our implementation of CPPN-NEAT makes use of 6 activation functions; Sine, Cosine, Identity, Gaussian, Absolute and Sigmoid. The first population begins unevolved with each member consisting of $3$ input neurons and a single output neuron before random initialisation. Network structure is recorded for all population members through the use of JSON files.
 
\subsubsection{Bezier}
Bezier splines are the direct encoding compared in our research. To produce a leg using the Bezier representation the genotype must contain the information of $5-10$ 3D Bezier splines; $3-8$ control points for each spline; and a thickness value for each spline with a value between $1-3$. The mapping of genotype to phenotype to produce a single leg can be seen in Figure \ref{Representation}(ii). 

To evolve the Bezier representation a genetic algorithm (GA) using tournament selection and two-point crossover is used. The algorithm starts with a population of 20 individuals randomly generated with values within the specified ranges. The evaluation of the current population through simulation in Project Chrono provides a fitness used to rank the individuals. Elitism is used to carry through the best parent into the next population.  The remaining 19 children are generated using selection and crossover. Tournament selection is applied to find the parents using $4$ randomly sampled individuals to find $1$ parent. The two crossover points are randomly chosen from the shortest genome of the chosen parents. 

All members of the child population are exposed to uniform variation of Bezier control points as part of a mutation step. Uniform variation takes the form of a Gaussian with a mean of $0$ and a standard deviation of $10\%$ of the maximum range of a control point. Further possible mutations alter the structure of each leg with a (i) $20\%$ chance of a random Bezier's thickness changing to within the limits (ii) $20\%$ chance of a control point from a random Bezier being added or removed within the limits (iii) $10\%$ chance of a random Bezier being removed or added with random initialisation within the limits.

\subsection{Experiment 1}

Due to the high likelihood of a CPPN producing legs that are not contiguously linked from the top plane to the bottom plane of the grid, two processes were investigated to ensure compliance. One approach uses an adaptive threshold to ensure compliance, the second approach (already reported in the literature \cite{Auerbach2014EnvironmentalMachines}) scales the largest artefact to meet the leg length constraint. 

\subsubsection{Thresholding}
\begin{figure*}[thpb]
  \centering
  \includegraphics[width=\textwidth]{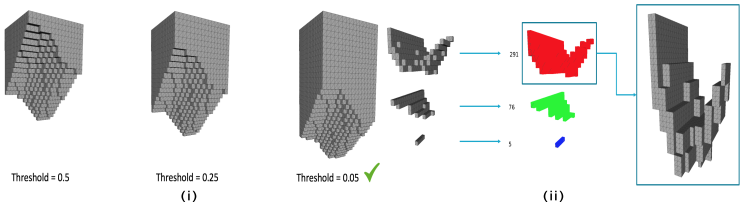}
  \caption{(i) The output of a CPPN given three decreasing thresholds. The threshold begins at $0.5$ giving a cell an equal chance of being ''full" or ''empty". If a leg does not satisfy the length constraint the threshold is lowered until a leg does satisfy the constraint.(ii) A CPPN generated leg that does not satisfy the length constraints is separated into segregated elements using Hopcroft's and Tarjan's connected components algorithm \cite{Hopcroft1973AlgorithmManipulation}. The largest element is computed with all smaller elements removed from the voxelised form. The largest element can then be rescaled to fill the voxel environment.}
  \label{ConstraintSatisfaction}
\end{figure*}

Adaptive thresholding adjusts the output threshold for the CPPN. Using a connected components algorithm by Hopcroft and Tarjan \cite{Hopcroft1973AlgorithmManipulation}, features are labeled with a complete leg comprising of a connected grid of voxels from top to bottom. If a connected leg does not exist the threshold deciding whether the numerical output of the CPPN is either to be ``full" or ``empty" is lowered to promote more full voxels. The leg is then re-evaluated with the cycle repeating until the voxelised space contains a feasible leg. Figure \ref{ConstraintSatisfaction}(i) demonstrates how the change in thresholds affects the appearance of a leg.

\subsubsection{Scaling}
The rescaling of occupied voxels has several steps that allow for an acceptable leg to be produced. First a leg is checked whether it is compliant using the same Hopcroft and Tarjan algorithm to label features. If a leg is not compliant then all artefacts but the largest are removed from the voxelised space. The size of the largest artefact is derived and the required scaling to ensure it stretches to the bounds of the grid is computed. Finally, the single artefact is scaled accordingly, the process follows Figure \ref{ConstraintSatisfaction}(ii).

\subsubsection{Results}
\begin{figure*}[thpb]
  \centering
  \includegraphics[width=\textwidth]{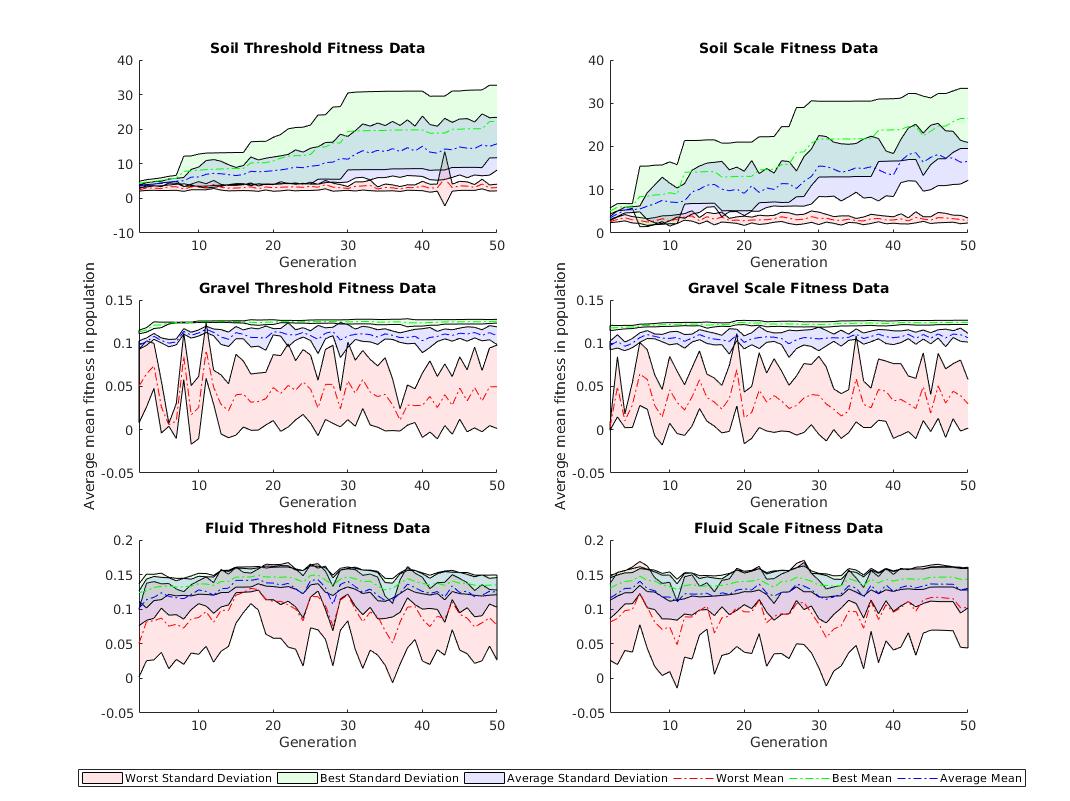}
  \caption{Graph comparing CPPN constraint satisfaction methods through fitness progression of $10$ experimental repeats across $50$ generations for (top-left) soil environment using the threshold method; (top-right) soil environment using the scale method; (middle-left) gravel environment using the threshold method; (middle-right) gravel environment using the scale method; (bottom-left) fluid environment using the threshold method; (bottom-right) fluid environment using the scale method. Green = best fitness, blue = mean fitness, red = worst fitness. Shaded areas denote standard error.}
  \label{CPPN_Acum}
\end{figure*}

Fitness graphs in Figure \ref{CPPN_Acum} highlight the similarity in performance between approaches used to generate compliant CPPN artefacts. The three environments, soil, gravel and fluid, all exhibit distinct characteristics which produce different fitness traits. Soil legs show a steady increase in the average best fitness across the $50$ generations. The best fitness achieved by the thresholding approach in soil was $29.0705$ with the average best of the $10$ runs being $20.1488$. Conversely, scaling the CPPN output produces an improved best average of $23.1240$ ($>20.1488$) but a decrease of the best fitness with a result of $29.0593$ ($<29.0705$).

The thresholding approach in gravel produces an evolved leg with a best fitness value of $0.1294$ with an average across all $10$ experiments of $0.1239$. Using the scaling method in gravel produces increases in best fitness and the best average with values of $0.1582$ and $0.1243$ respectively. CPPN outputs for legs within the fluid environment see the same best fitness value of $0.1560$ for multiple legs created using either of the two constraining methods. The scaling method does produce a better average best over thresholding with values of $0.1433$ and $0.1380$ respectively. 

Conclusions about the simulation environments can be drawn from the fitness plots in Figure \ref{CPPN_Acum}. Gravel and fluid environments present a distinguished difference to the soil environment with very little increase in the fitness of members after $\approx10$ generations. Furthermore, in comparison to the soil environment with a maximum fitness of $29.0705$ the gravel and fluid environments share significantly lower fitness values. The lower fitness is not unexpected and can be attributed to the higher torques required to complete a ``step" in particle environments.

The standard deviation of fitnesses in gravel for both the best and average plots is very close to converging with the best and average mean lines. The concurrence of both a low standard deviation and early increase in fitness followed by a visually steady plateau leads to the inference that gravel is an unchallenging environment. 

Results comparing CPPN constraint methods are not statistically significant and for that reason our first research question remains undetermined. Scaling the output is more complex than thresholding but yields increased values for best fitness across the ten experimental repeats. Therefore, further comparisons between direct and indirect representations will assume use of the scaling method.

% Evident through the increased fitness values shared by almost all scaled legs and increased average best of all scaled legs, scaling is seen as a better solution to constraining CPPN outputs. Scaling the output is more complex than using an adaptive threshold but the increased performance warrants its use. This answers our first research question with scaling being the preferred method of the two, therefore comparisons .

\subsection{Experiment 2}

\subsubsection{Results}
Comparisons between indirect and direct representations are statistically significant for all environments with p-values of $0.02574$ for soil, $0.00278$ for gravel and $0.00018$ for fluid. 

\begin{figure}
  \centering
  \includegraphics[width=\linewidth]{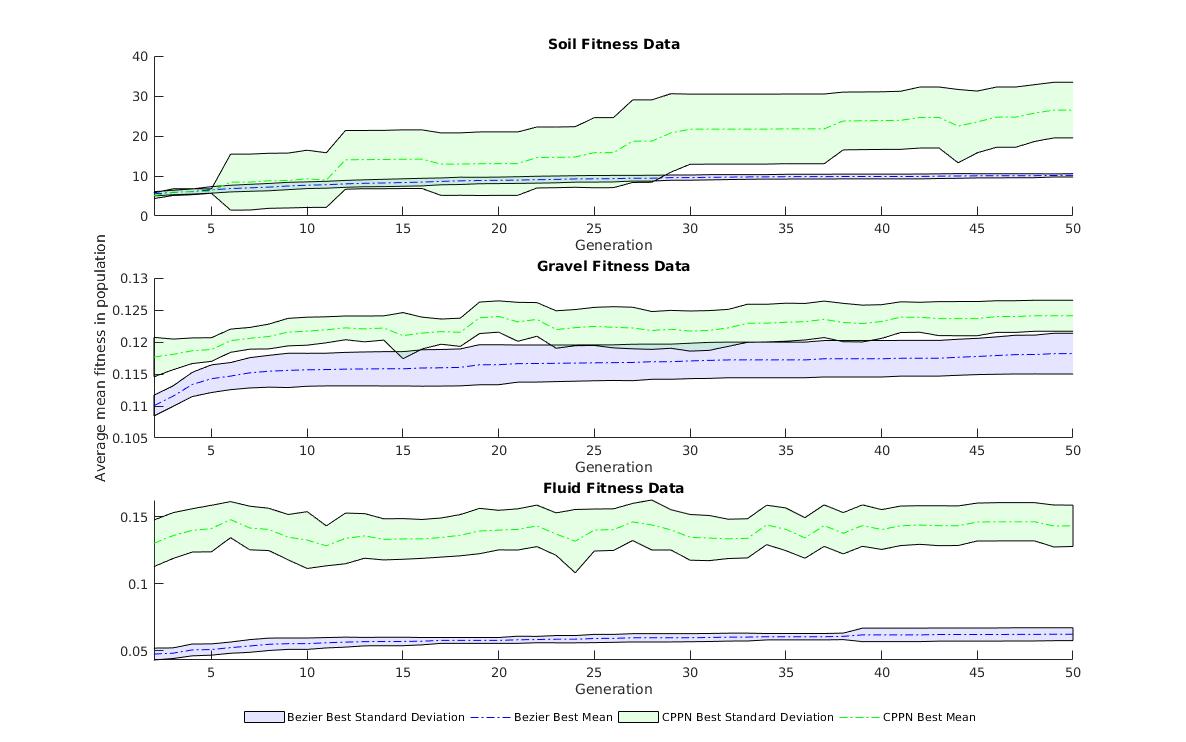}
  \caption{Graph comparing direct and indirect representations through fitness progression across $50$ generations for (top) soil environment; (centre) gravel environment; and (bottom) fluid environment. Green = CPPN-NEAT, blue = Bezier. Shaded areas denote standard error.}
  \label{Rep_Accum}
\end{figure}

Fitness graphs comparing the direct and indirect representations for all 3 environments can be found in Figure \ref{Rep_Accum}. Evident when comparing Bezier fitness to CPPN fitness is the great variation in standard deviation experienced by CPPN fitnesses. This is due to the inherent nature of indirect encodings where small changes in genotype can correlate to large changes in phenotype and as an extension fitness. This agrees with a wealth of literature \cite{Hornby2002CreatingEvolution}.

In soil the Bezier representation was able to produce a leg with a fitness of $11.0619$, a value which is dwarfed by the CPPN value of $29.0593$. The massive increase in fitness by CPNNs is shared by the fluid environment with Bezier's achieving a value of $0.0748$ whilst CPPNs achieved a value of $0.1479$. The closest Beziers came to CPPNS  was in the gravel environment where fitness values of $0.1237$ and $ 0.1404$ were recorded respectively. 

It is evident from Figure \ref{Rep_Accum} that the indirect representation of CPPN-NEAT yields greater increase of fitness over the direct Bezier representation. Therefore, in our context of evolving environmentally-specific hexapod tibia the indirect representation of CPPN-NEAT is better than the indirect representation of Bezier splines. The answer to our second research question is satisfied with this result.

\subsection{Morphology}

%Evolvability and how soil was able to exploit the representation to get to the best possible fitness. For comparatively difficult environments to soil biologically formations can be found. The fluid environment generated both blocky legs and streamlined legs contrary to bezier formations, 

\begin{figure}
  \centering
  \includegraphics[width=8cm]{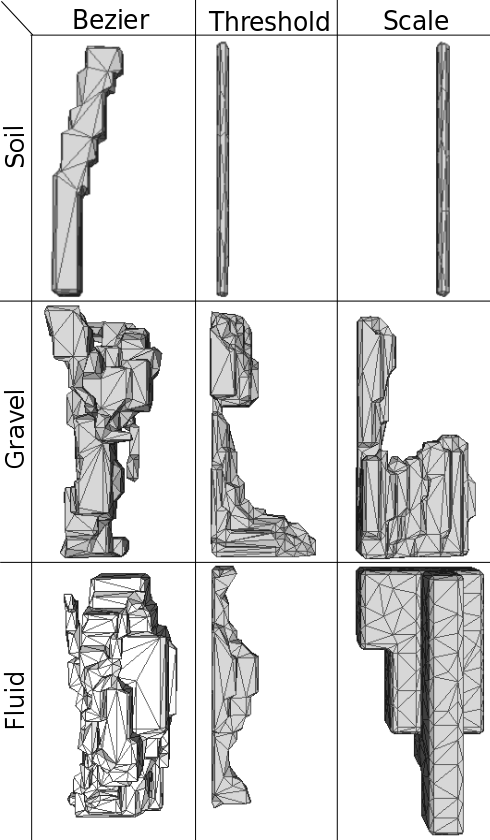}
  \caption{.STL files of the fittest leg for each environment type (soil, gravel and fluid) and creation method (Bezi er Representation using a Genetic Algorithm, CPPN-NEAT with adjustable threshold for constraint satisfaction and CPPN-NEAT with scaled artefact for constraint satisfaction). Note the disparity in resolution between fluid legs evolved using CPPNs with the streamlined leg expressed using higher resolution.}
  \label{OptimalLegs}
\end{figure}

The most optimal legs evolved for each environment and representation are pictured in Figure \ref{OptimalLegs}. The morphologies of the soil legs are similar in that they gravitate towards producing legs that require less material. This shows the impact the second term of the fitness function has on reducing excess material. The first term of the fitness function, the reduction of torques, is only possible to an extent in this simple environment. The exploitation of the design space by CPPN-NEAT is on display as generated legs for soil still satisfy length constraints but produce fitness values that are likely converging on the global optima for the current fitness function. The small difference in fitness values of the CPPN soil legs (i.e. $29.0705$ and $29.0593$) is likely due to the positioning of the thin leg; in the thresholding method it sits to the back of the design space whereas in the scale method its sits closer to the front.

The morphology of gravel legs tend towards shapes with large foot tips as this allows them to distribute the force of the ``step" preventing it from sinking too far below top layer of gravel. Additionally, the streamlined shapes allow the leg to push forward through particles of gravel. The biological mimicking properties of CPPNs and the complexity that NEAT affords is displayed through the results within the gravel environment. In particular, the leg evolved using CPPN-NEAT and adaptive thresholding looks very much like our own evolved foot and ankle. The gravel leg evolved using CPPN-NEAT and the scaling method has the same characteristics but has a slightly different approach. It still has a pillar which allows it to satisfy the length constraint; a large surface area on the base to distribute weight; and a foot that is streamline but the top of which is too high for particles to sit on top of. 

Fluid legs evolved using CPPNs share a common best fitness value (0.1560), irrespective of constraining method. Figure \ref{OptimalLegs} shows the contrasting morphology of two of the high fitness legs evolved for the fluid environment. Evolution finds two techniques to generate high fitness legs in fluid, firstly small cross-sectional area commonly associated with hydrodynamic designs, and secondly large, flat faces perpendicular to motion to deflect water with minimal drag. 

% The fluid environment is the most computationally intensive and certainly the hardest to replicate real-world fluid dynamics.

\section{Conclusion}

We compare two representations for defining morphologies in a voxelised space and evaluate their ability to produce evolved artefacts. The first representation is a direct encoding composed of 3D Bezier splines, the second is an indirect encoding utilising the current state-of-the-art CPPN-NEAT. Both representations are used to evolve environment-specific legs for a hexapod robot in three high-fidelity simulation environments. The evolved morphologies and recorded statistics are directly compared to determine which representation is best able to exploit the design space to generate high-fitness legs.

Translating CPPN-NEAT from genotype to phenotype can result in unrealistic artefacts, so our first experiment compares two methods of rectifying the output. The first utilises an adaptive threshold on the output neuron of the CPPN to create more occupied voxels, the second scales the largest produced artefact within the design space. Scaling the output is the preferred method of constraint satisfaction and was the method used when comparing direct and indirect representations in experiment 2. 

% What has been learnt?
Our hypothesis that \emph{indirect representations allow for greater exploitation of the environment, in turn gaining a higher fitness} is proven to be correct in experiment 2. Quantitative data supports the hypothesis with considerable fitness increases by our indirect representation for all environments. The morphologies of evolved legs prove that CPPN-NEAT as a representation is better able to exploit the design space by producing biological mimicking artefacts and designs that the Bezier representation was unable to produce.
  
Our approach evolves the components of a hexapod most likely to benefit from environmental specialisation due to direct environmental interaction. Evolved components are easily instantiated using standard 3D printable files and can be fixed to one of our hexapods for increased performance on its next mission. Extension of this work will investigate the transfer of these simulated artefacts into the real-world and the improvement of our simulated environments.

\appendix
\linespread{0.9}
\section*{Appendix A} \label{Appendix A}

\subsection{Simulation}
\noindent
Gravel Particle Size: $0.01m$\\
No. Gravel Particles: $380$\\
Fluid particles size: $0.016m$\\
No. Fluid Particles: $316$

\subsection{CPPN-NEAT}
\noindent
\emph{Probabilities sampled from uniform distribution except mutation of connection weights which is sampled from a normal distribution.}\\
CPPN Voxel Threshold: $0.5$ (Unless using adaptive threshold)\\
CPPN Functions: Sine, Cosine, Identity, Guassian, Absolute, Sigmoid\\
Elitism: $1$ member from parent population\\
Offspring without Crossover: $p=0.25$\\
Interspecies cross-over: $0.05$\\
Add Node: $p=0.1$\\
Add Connection: $p=0.25$\\
Mutate Connection Weight: $p=0.25$\\
Mutate Node Activation: $p=0.1$

\subsection{Bezier-GA}
\noindent
\emph{Values are inclusive. Probabilities sampled from uniform distribution.}\\
Minimum No. of Bezier: $5$\\
Maximum No. of Bezier: $10$\\
Minimum Control Points: $3$\\
Maximum Control Points: $8$\\
Minimum thickness of Bezier: $1$ Voxel\\
Maximum Thickness of Bezier: $3$ Voxels\\
Genetic Algorithm Elitism: $1$ Member from parent population\\
Tournament Size: $4$\\
Control Point Mutation: $x\times\mathcal{N}(0,(16\times0.1)/4)$\\
$y\times\mathcal{N}(0,(32\times0.1)/4)$\\
$z*\mathcal{N}(0,(16\times0.1)/4)$\\
Mutate No. of Bezier: $p=0.1$, $p=0.5$ add/remove\\
Mutate No. of Control Points $p=0.2$, $p=0.5$ add/remove\\
Mutate Thickness of Bezier $p=0.2$

\bibliographystyle{ieeetr}

% \bibliography{References.bib}

\end{document}